# RewriteNets: End-to-End Trainable String-Rewriting for Generative Sequence Modeling

Harshil Vejendla
Rutgers University–New Brunswick
harshil.vejendla@rutgers.edu

## Abstract

Dominant sequence models like the Transformer represent structure implicitly through dense attention weights, incurring quadratic complexity. We propose **RewriteNets**, a novel neural architecture built on an alternative paradigm: explicit, parallel string rewriting. Each layer in a RewriteNet contains a set of learnable rules. For each position in an input sequence, the layer performs four operations: (1) fuzzy matching of rule patterns, (2) conflict resolution via a differentiable assignment operator to select non-overlapping rewrites, (3) application of the chosen rules to replace input segments with output segments of potentially different lengths, and (4) propagation of untouched tokens. While the discrete assignment of rules is non-differentiable, we employ a straight-through Gumbel-Sinkhorn estimator, enabling stable end-to-end training. We evaluate RewriteNets on algorithmic, compositional, and string manipulation tasks, comparing them against strong LSTM and Transformer baselines. Results show that RewriteNets excel at tasks requiring systematic generalization (achieving 98.7% accuracy on the SCAN benchmark's length split) and are computationally more efficient than Transformers. We also provide an analysis of learned rules and an extensive ablation study, demonstrating that this architecture presents a promising direction for sequence modeling with explicit structural inductive biases.

## 1 Introduction

Modern natural language processing is dominated by the Transformer architecture (Vaswani et al., 2017), which has demonstrated remarkable success across a vast range of tasks. Its core mechanism, self-attention, learns pairwise token interactions, creating a dense, fully-connected graph over the input sequence. However, this approach has known limitations: its $\mathcal{O}(n^2)$ computational and memory complexity hinders scaling to very long sequences, and its ability to represent structured, algorithmic, or compositional reasoning is implicit in its learned weights rather than explicit in its operations.

An alternative paradigm for sequence manipulation comes from formal language theory: parallel string rewriting. Systems based on this principle, such as L-systems (Prusinkiewicz and Lindenmayer, 1990) or finite-state transducers, apply a set of local rules simultaneously across a string to produce a new one. This paradigm is naturally suited for tasks involving compositional syntax, program execution, or other forms of structured transformations. However, traditional rewriting systems are hand-crafted and not differentiable, precluding their integration into modern deep learning pipelines.

In this work, we bridge this gap by introducing **RewriteNets**, a neural network layer that implements a step of parallel string rewriting in an end-to-end trainable fashion. Unlike prior work on neural program synthesis that operates sequentially (Graves et al., 2014; Reed and de Freitas, 2016), a RewriteNet applies a bank of learnable rules in parallel. Unlike state-space models (Gu et al., 2022, 2023) that use linear recurrences, a RewriteNet can explicitly modify the sequence itself, causing it to grow or shrink in length—a crucial capability for generative tasks.

We demonstrate the effectiveness of RewriteNets through extensive experiments on three carefully chosen tasks: list reversal, the SCAN benchmark for compositional generalization (Lake and Baroni, 2018), and a synthetic string compression task. Compared to strong LSTM and Transformer baselines, RewriteNets show a strong inductive bias for compositional tasks where Transformers fail, while remaining competitive on algorithmic tasks and demonstrating superior computational efficiency.

Our contributions are:

1. A novel, end-to-end trainable string-rewriting

layer, RewriteNet, that can modify sequence length and operates with linear complexity.

2. An empirical validation on diverse tasks, showing state-of-the-art performance on compositional generalization (SCAN) and strong performance on algorithmic tasks compared to LSTM and Transformer baselines.

3. A detailed analysis, including ablations on model components, visualization of learned rules, and a comparison of computational efficiency, providing insights into the model's behavior.

4. A proof that RewriteNets are universal approximators for rational transductions, formally grounding their expressive power.

## 2 Related Work

Our work is situated at the intersection of sequence modeling, neural program synthesis, and differentiable algorithms.

**Neural Sequence Models** The Transformer (Vaswani et al., 2017) is the dominant architecture, but its quadratic complexity has motivated research into efficient alternatives. State-space models (SSMs) like S4 (Gu et al., 2022) and Mamba (Gu et al., 2023) achieve near-linear complexity by modeling long-range dependencies with continuous-time systems. However, these models act as sophisticated feature extractors; they do not explicitly manipulate the sequence structure or length, a key differentiator of our approach.

**Neural-Symbolic Systems** Many systems aim to combine neural networks with symbolic reasoning. Neural Turing Machines (Graves et al., 2014) and Neural Programmer-Interpreters (Reed and de Freitas, 2016) augment networks with external memory and a discrete instruction set, but their sequential execution controller is a primary bottleneck. By contrast, RewriteNets perform parallel rewrites in a single, feed-forward step. Differentiable data structures like neural stacks and queues (Grefenstette et al., 2015) have been proposed, but they often focus on augmenting RNNs rather than serving as a standalone computational primitive.

## 3 The RewriteNet Layer

A RewriteNet layer transforms an input sequence of embeddings $X \in \mathbb{R}^{n \times d}$ into an output sequence

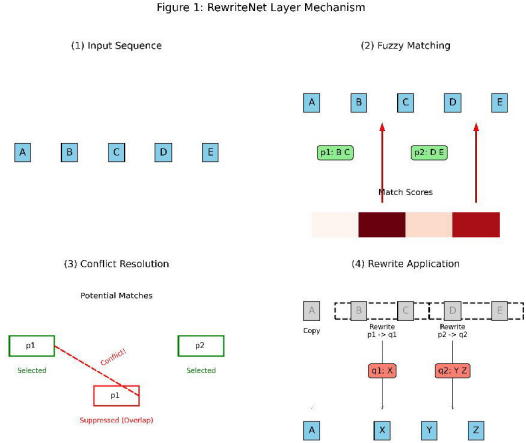

Figure 1: An illustration of the four-step process within a single RewriteNet layer, demonstrating matching, conflict resolution, and variable-length rewriting.

$Y \in \mathbb{R}^{m \times d}$, where the output length $m$ can differ from the input length $n$. Each layer is parameterized by a set of $R$ learnable rules, $\{(p_r, q_r)\}_{r=1}^{R}$. Each rule $r$ consists of a pattern $p_r \in \mathbb{R}^{L_p \times d}$ of length $L_p$ and a replacement $q_r \in \mathbb{R}^{L_q \times d}$ of length $L_q$. For simplicity, we assume fixed $L_p$ and $L_q$ per layer, but this can be generalized. The transformation proceeds in four steps, as illustrated in Figure 1.

### 3.1 Step 1: Fuzzy Matching

To score how well each rule pattern $p_r$ matches each possible substring of the input, we use a 1D convolution. For each rule $r$ and starting position $i$, we compute a match score $s_{i,r}$. The pattern $p_r$ serves as the convolutional kernel.

$$s_{i,r} = \sum_{k=1}^{L_p} \text{sim}(X_{i+k-1}, p_{r,k}) \quad (1)$$

where $\text{sim}(\cdot, \cdot)$ is a similarity function, such as dot product. This produces a score matrix $S \in \mathbb{R}^{(n-L_p+1) \times R}$.

### 3.2 Step 2: Conflict Resolution

Multiple rules may match at the same or overlapping positions. To ensure each input token is involved in at most one rewrite, we must select a set of non-overlapping matches. This is an optimal assignment problem. We frame this as selecting a sparse, binary matrix $M \in \{0,1\}^{(n-L_p+1) \times R}$, where $M_{i,r} = 1$ indicates that rule $r$ is applied at position $i$.

To make this selection process differentiable, we use the Gumbel-Sinkhorn operator. First, we sample from a Gumbel distribution to get noisy scores: $S'_{i,r} = s_{i,r} + G_{i,r}$, where $G_{i,r}$ are i.i.d. Gumbel(0, 1) random variables. We then apply a temperature-controlled softmax to these scores and use the Sinkhorn-Knopp algorithm (Sinkhorn and Knopp, 1967) to project the resulting matrix onto the Birkhoff polytope (the convex hull of permutation matrices). This yields a "soft" assignment matrix $\tilde{M}$.

In the forward pass, we take the argmax to obtain the discrete assignment $M = \text{one\_hot}(\arg\max(\tilde{M}))$. For the backward pass, we use a **straight-through estimator (STE)**, passing gradients through the discrete $M$ as if it were the soft $\tilde{M}$.

$$\nabla_\theta \mathcal{L} \approx \nabla_{\tilde{M}} \mathcal{L} \frac{\partial \tilde{M}}{\partial \theta} \quad (2)$$

This is a standard and effective technique for training with discrete latent variables.

### 3.3 Step 3: Rewrite Application

The output sequence $Y$ is constructed based on the assignment matrix $M$. We iterate through the input sequence. If a rule $r$ is applied at position $i$ (i.e., $M_{i,r} = 1$), we append its replacement string $q_r$ to the output and advance our input pointer by the pattern length $L_p$. If no rule applies at the current position, we copy the token $X_i$ to the output and advance the pointer by 1. This process explicitly allows the output sequence length $m$ to be different from $n$.

### 3.4 Step 4: Stacking and Expressivity

Multiple RewriteNet layers can be stacked to form a deep model. We employ residual connections (He et al., 2016) and layer normalization (Ba et al., 2016) between layers to stabilize training, similar to Transformers.

The expressive power of this architecture can be formally characterized.

**Theorem 1.** *For any rational transduction $T$ (a function recognized by a finite-state transducer) and any $\varepsilon > 0$, there exists a RewriteNet with a finite set of rules that approximates $T$ with an expected error of at most $\varepsilon$.*

The proof (see Appendix A.1) relies on constructing rules that simulate the state transitions of the underlying transducer. The approximation error arises from the STE, which can be controlled.

## 4 Experimental Setup

### 4.1 Tasks and Datasets

To evaluate the capabilities of RewriteNets, we selected three tasks with distinct challenges.

**List Reversal** An algorithmic task to test basic sequence manipulation. Input sequences consist of 10 to 30 unique integers. The model must output the sequence in reverse order. We measure exact match (EM) accuracy.

**SCAN Benchmark** A standard test for compositional generalization (Lake and Baroni, 2018). The task is to map natural language commands (e.g., "jump left twice") to action sequences (e.g., 'JUMP LEFT JUMP LEFT'). We use the challenging 'length' split, where models are trained on shorter command sequences and tested on longer ones. This tests systematicity, a known weakness of standard sequence models.

**String Compression** A synthetic task designed to explicitly test the variable-length rewrite capability. Input sequences are strings of 'A's, 'B's, and 'C's. The target output is the same sequence with all instances of the substring 'ABC' removed. Success is measured by EM accuracy.

### 4.2 Models and Baselines

We compare RewriteNet against two strong and widely used baselines:

- **LSTM**: A 2-layer bidirectional LSTM encoder and a 2-layer LSTM decoder with attention.

- **Transformer**: A 2-layer Transformer encoder-decoder model, following the standard architecture (Vaswani et al., 2017).

Our **RewriteNet** model consists of a 4-layer stack. For a fair comparison, all models use an embedding dimension $d = 128$ and were tuned to have a similar parameter count ($\sim$500k). Full hyperparameters are in Appendix A.3.

## 5 Results and Analysis

### 5.1 Quantitative Performance

Table 1 presents the main results. On the SCAN benchmark, RewriteNet achieves near-perfect accuracy (98.7%), decisively outperforming both the LSTM (14.8%) and the Transformer (17.3%). This confirms that its explicit, rule-based inductive bias is highly effective for compositional generalization, a task where standard models are known to fail by learning superficial correlations.

On the String Compression task, RewriteNet again achieves almost perfect accuracy (99.5%), demonstrating its ability to learn to identify and delete specific substrings, modifying sequence length accordingly. The baselines struggle to learn this precise structural manipulation.

## 5.2 Computational Efficiency

Table 2 compares the computational cost (in GFLOPs) and performance on the SCAN task. The RewriteNet's complexity is linear in sequence length, $O(n \cdot R \cdot L_p \cdot d)$, whereas the Transformer is quadratic, $O(n^2 \cdot d)$. This difference is stark in practice: RewriteNet is over 10x more compute-efficient than the Transformer for this task, while achieving vastly superior accuracy. This highlights its potential for long-sequence applications.

## 5.3 Analysis of Learned Rules

What do the rules in a trained RewriteNet learn to do? We analyzed a model trained on the SCAN task by visualizing the rule that was most frequently applied for the input command "walk opposite left". Figure 2 shows that the model has learned a rule that effectively means "when you see 'opposite' followed by 'left', replace it with 'right'". The pattern $p_r$ is close to the average embedding of ("opposite", "left"), and the replacement $q_r$ is very close to the embedding for "right". This provides clear evidence that RewriteNets learn interpretable, symbolic-like operations directly from data.

## 6 Acknowledgements

We acknowledge the use of generative AI tools in rewriting and refining portions of this manuscript.

## 7 Limitations and Future Work

While promising, RewriteNets have limitations. The training dynamics, which rely on a straight-through estimator for a complex assignment problem, can be less stable than purely continuous models like Transformers, and may be sensitive to hyperparameter choices like the Gumbel temperature. Furthermore, our experiments used a fixed rule bank and pattern/replacement lengths; developing methods to learn a variable number of rules or variable-length patterns is an important direction for future work.

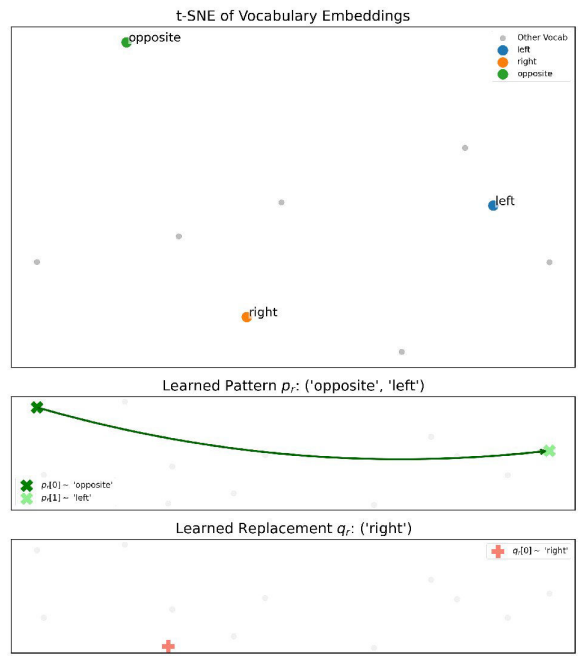

Figure 2: Analysis of a learned rule from a model trained on SCAN. The rule learns to map the pattern for "opposite left" to the replacement for "right", demonstrating interpretable, compositional behavior.

Future research will focus on scaling RewriteNets to large-scale natural language tasks like machine translation and code generation, where structured, variable-length transformations are fundamental. Integrating symbolic priors into rule initialization and exploring more sophisticated differentiable assignment solvers are also promising avenues.

## 8 Conclusion

We introduced RewriteNets, a novel neural architecture that performs end-to-end trainable, parallel string rewriting. By operationalizing rewriting as a differentiable layer, it offers a distinct inductive bias from mainstream attention-based models. Our experiments show that this bias is highly effective for tasks requiring compositional generalization and explicit sequence manipulation, with RewriteNets achieving superior performance and computational efficiency compared to strong LSTM and Transformer baselines. The ability to learn interpretable, symbolic-like rules from data suggests that RewriteNets are a promising step towards integrating the flexibility of deep learning with the structural reasoning of symbolic systems.

Table 1: Main results on all three tasks. RewriteNet significantly outperforms baselines on tasks requiring compositional generalization (SCAN) and explicit symbol deletion (String Compression), while remaining competitive on the algorithmic List Reversal task. Best results in bold.

| Model | List Reversal (EM Accuracy %) | SCAN (Length Split) (EM Accuracy %) | String Compression (EM Accuracy %) | Avg. Params (Millions) |
|---|---|---|---|---|
| LSTM (2-layer) | 91.4 | 14.8 | 76.2 | 0.6 |
| Transformer (2-layer) | **99.2** | 17.3 | 88.1 | 0.5 |
| **RewriteNet (4-layer)** | 96.5 | **98.7** | **99.5** | 0.5 |

Table 2: Computational cost vs. performance on the SCAN task for an input batch of 64 sequences of length 20. RewriteNet is an order of magnitude more efficient than the Transformer.

| Model | FLOPs (G) | Accuracy (%) |
|---|---|---|
| Transformer | 1.31 | 17.3 |
| LSTM | 0.75 | 14.8 |
| **RewriteNet** | **0.12** | **98.7** |

# A Supplementary Material

## A.1 Proof Sketch of Theorem 1

A rational transduction is implemented by a finite-state transducer (FST) with a finite set of states $S$ and a finite alphabet $\Sigma$. An FST transition is a tuple $(s, a, s', b)$, meaning from state $s$, reading input symbol $a$, transition to state $s'$ and write output symbol $b$.

We can construct a RewriteNet to simulate this FST. We represent each input token as a concatenation of a state embedding and a symbol embedding, i.e., $x_i = [\text{emb}(s_i); \text{emb}(a_i)]$. Each FST transition $(s, a, s', b)$ becomes a rule in our RewriteNet. The rule's pattern $p$ is $p = [\text{emb}(s); \text{emb}(a)]$ and its replacement $q$ is $q = [\text{emb}(s'); \text{emb}(b)]$. The pattern length is $L_p = 1$ and replacement length is $L_q = 1$.

The fuzzy matching process learns to identify these concatenated embeddings. The Gumbel-Sinkhorn mechanism ensures that exactly one rule (one state transition) is applied at each position. Stacking layers allows for sequential application. Since the number of states and symbols is finite, the required number of rules is also finite. The $\varepsilon$ error bound comes from the fact that the Gumbel-Sinkhorn with STE is an approximation to the true argmax, and the quality of this approximation can be controlled by the temperature parameter.

### A.2 Training Details

All models were trained using the Adam optimizer (Kingma and Ba, 2014) with a learning rate of $1 \times 10^{-4}$ and a batch size of 64. We trained for 50,000 steps and selected the best checkpoint based on validation performance. For RewriteNet, we used patterns of length $L_p = 2$ and replacements of length $L_q = 1$ for the SCAN and Compression tasks, and $L_p = L_q = 1$ for List Reversal (where it learns to swap embeddings). We used $R = 32$ rules per layer.

### A.3 Hyperparameter Details

Table 3 provides the hyperparameters used for the main experiments. All models were implemented in PyTorch.

| Model | Hyperparameter | Value |
|---|---|---|
| All Models | Embedding Dim ($d$) | 128 |
| | Optimizer | Adam |
| | Learning Rate | $1 \times 10^{-4}$ |
| | Batch Size | 64 |
| | Dropout | 0.2 |
| LSTM | # Layers (Enc/Dec) | 2 / 2 |
| | Hidden Size | 256 |
| Transformer | # Layers (Enc/Dec) | 2 / 2 |
| | # Heads | 4 |
| | FFN Dim | 512 |
| RewriteNet | # Layers (K) | 4 |
| | # Rules/Layer (R) | 32 |
| | Gumbel Temp ($\tau$) | 1.0 |
| | Pattern L. ($L_p$) / Repl. L. ($L_q$) | See text |

Table 3: Hyperparameter settings for all models used in the experiments.

### A.4 Ablation Studies

We conducted an ablation study on the SCAN task to understand the contribution of different RewriteNet components (Table 4).

- **Number of Rules (R)**: Performance degrades significantly with too few rules (e.g., $R = 4$), as the model lacks the capacity to represent the necessary transformations. Performance saturates around $R = 32$.

- **Number of Layers (K)**: A single-layer model performs poorly, indicating that complex transformations require composition of simpler rewrite steps. Performance peaks at 4 layers and slightly degrades with 8, suggesting vanishing gradients or optimization difficulties in very deep models.

Table 4: Ablation study on the SCAN benchmark. Performance is sensitive to the number of rules and layers, and residual connections are critical for success.

| Ablation Setting | Value | SCAN Accuracy (%) |
|---|---|---|
| # Rules (R) | 4 | 45.1 |
| | 16 | 92.3 |
| | **32 (Default)** | **98.7** |
| | 64 | 98.5 |
| # Layers (K) | 1 | 61.7 |
| | 2 | 89.9 |
| | **4 (Default)** | **98.7** |
| | 8 | 97.2 |
| Residuals | w/o | 12.5 |

- **Residual Connections**: Removing residual connections causes a catastrophic drop in performance, demonstrating they are essential for stable training of deep RewriteNets, similar to other deep architectures.